\documentclass[11pt]{article}
\usepackage{arabtex}
\usepackage{utf8}
\setcode{utf8}

\usepackage[final]{acl}

\usepackage{times}
\usepackage{latexsym}

\usepackage[T1]{fontenc}

\usepackage{microtype}

\usepackage{inconsolata}

\usepackage{graphicx}

\usepackage{tabularx}

\usepackage{subcaption}

\title{Abjad AI at NADI 2025: CATT-Whisper: Multimodal Diacritic Restoration Using Text and Speech Representations}
\author{Ahmad Ghannam, Naif Alharthi, Faris Alasmary, Kholood Al Tabash, \\
\textbf{Shouq Sadah, and Lahouari Ghouti} \\
        Abjad AI. King Khaled Road. Riyadh 11000, Saudi Arabia. \\ \texttt{\{aghannam,nalharthi,falasmary,kaltabash,ssadah,lghouti\}@abjad.com.sa}}

\begin{document}
\maketitle

\begin{abstract}
In this work, we tackle the Diacritic Restoration (DR) task for Arabic dialectal sentences using a multimodal approach that combines both textual and speech information. We propose a model that represents the text modality using an encoder extracted from our own pre-trained model named CATT. The speech component is handled by the encoder module of the OpenAI Whisper base model. Our solution is designed following two integration strategies. The former consists of fusing the speech tokens with the input at an early stage, where the 1500 frames of the audio segment are averaged over 10 consecutive frames, resulting in 150 speech tokens. To ensure embedding compatibility, these averaged tokens are processed through a linear projection layer prior to merging them with the text tokens. Contextual encoding is guaranteed by the CATT encoder module. The latter strategy relies on cross-attention, where text and speech embeddings are fused. The cross-attention output is then fed to the CATT classification head for token-level diacritic prediction. To further improve model robustness, we randomly deactivate the speech input during training, allowing the model to perform well with or without speech. Our experiments show that the proposed approach achieves a word error rate (WER) of 0.25 and a character error rate (CER) of 0.9 on the development set. On the test set, our model achieved WER and CER scores of 0.55 and 0.13, respectively.
\end{abstract}

\section{Introduction}

Diacritics are essential for accurate interpretation, pronunciation, and meaning in Arabic. However, in most informal writing such as social media, messaging, or transcribed speech they are omitted. While native speakers often infer the intended forms from context, the absence of diacritics introduces significant ambiguity, particularly in dialects where phonetic and morphological variation is high and orthographic conventions are inconsistent. This not only challenges human readers but also degrades the performance of downstream NLP tasks such as speech synthesis, machine translation, and information retrieval. The NADI 2025 shared task overview \cite{nadi2025} highlights that DR remains particularly difficult for dialectal Arabic due to limited annotated data, regional variability, and inconsistent spelling practices.  Traditional DR approaches rely solely on text, ranging from rule-based systems and n-gram models to transformer-based language models such as BERT \cite{devlin2019bertpretrainingdeepbidirectional}. These methods often fail when orthographic cues alone are insufficient, an issue exacerbated in dialectal and code-switched text. In contrast, speech carries prosodic and phonetic signals that can directly disambiguate diacritic placement, offering a valuable complement to text.  


In this work, we propose CATT-Whisper, a multimodal DR system that integrates a CATT \cite{alasmary2024catt} text encoder with the Whisper \cite{pmlr-v202-radford23a} speech encoder. We evaluated two fusion strategies: \textbf{(i) Early fusion}: projected speech embeddings are merged with text embeddings before passing them to CATT encoder as inputs. \textbf{(ii) Cross-attention fusion}: the output of the CATT encoder is fused with the speech embeddings from Whisper using cross attention layer, followed by the classification layer. 

Our contributions are: (i) A multimodal DR system for Arabic dialects combining large-scale pre-trained text and speech encoders. (ii) Comparative analysis of early fusion vs. Cross-attention fusion. (iii) A modality-robust training scheme for variable speech availability. Our full codebase, including pre-trained models and training scripts, is publicly available \footnote{\url{https://github.com/abjadai/catt-whisper}}, ensuring reproducibility and facilitating further research in multimodal DR.

\section{Background}
\subsection{Task Setup}
The DR shared subtask at NADI 2025 focuses on restoring missing diacritics in Arabic text, with the option to also use speech for better performance. Unlike most previous work that only targets MSA, this task also covers Classical Arabic, dialects, and code-switched text, which are more challenging.
Some examples are provided in Table \ref{tab:examples}.
\begin{table}[h]
\centering
\begin{tabularx}{\linewidth}{|l|>{\centering\arraybackslash}X|}
\hline
Example Input 1         &            \<عندكو شوربة ايه النهرده>      \\ \hline
CATT                    &            \<عِنْدَكُو شُورْبَةُ ايه النَّهْرَدَهْ>    \\  
CATT-Whisper            &            \<عَندُكُو شوربِة اِيه النِهَردَه>     \\ 
\hline
Reference               &            \<عَندُكُو شوربِة اِيه النِهَردَه>     \\  
\hline \hline
Example Input 2         &            \<عايز شوية وأت لتجهيز الاكل>      \\  \hline
CATT                    &            \<عَايَزَ شُوِيَّةً وَأْتْ لِتَجْهِيزِ الاكْلِ>      \\  
CATT-Whisper            &            \<عَايِز شوَيَّة وَأت لِتَجهِيز الاَكل>      \\  
\hline
Reference               &            \<عَايِز شوَيَّة وَأت لِتَجهِيز الاَكل>      \\  
\hline
\end{tabularx}
\caption{
Examples from the NADI 2025 Subtask 3 dataset (dev/test). CATT (text-only) and CATT-Whisper (speech-enhanced) outputs compared with references, showing how speech features resolve phonological ambiguities.
}
\label{tab:examples}
\end{table}

\subsection{Dataset}
Our experiments were conducted using the NADI 2025 DR dataset, provided as part of the shared task, which is publicly available on Hugging Face \footnote{\url{https://huggingface.co/datasets/MBZUAI/NADI-2025-Sub-task-3-all}}. The dataset covers a mix of dialectal, multi-dialectal, and Classical Arabic varieties, with some segments exhibiting code-switching between Arabic and other languages. The dataset is a combined collection derived from several resources, namely MDASPC \cite{6487288}, TunSwitch \cite{abdallah2023leveragingdatacollectionunsupervised}, ArzEn \cite{hamed-etal-2020-arzen}, Mixat \cite{al-ali-aldarmaki-2024-mixat}, ClArTTS \cite{kulkarni2023clartts}, and ArVoice \cite{toyin2025arvoicemultispeakerdatasetarabic}.
While the CATT and Whisper models we use in our system were already pretrained on their respective large-scale corpora, the NADI 2025 DR dataset used exclusively for fine-tuning the combined architecture for this DR task.
The provided training data consists of multiple sub-datasets, summarized in Table~\ref{tab:dataset}.
\begin{table}[h]
\centering
\begin{tabular}{l l c r }
\hline
\textbf{Dataset} & \textbf{Type} & \textbf{Dia.} & \textbf{Train} \\ \hline
MDASPC    & Multi-dialectal  & True  & 60,677 \\ 
TunSwitch & Dialectal, CS    & True  & 5,212  \\ 
ArzEn     & Dialectal, CS    & False & 3,344  \\ 
Mixat     & Dialectal, CS    & False & 3,721  \\ 
ClArTTS   & CA & True  & 9,500  \\ 
ArVoice   & MSA              & True  & 2,507  \\ \hline
\end{tabular}
\caption{Statistics of the NADI 2025 Subtask 3 datasets. CA = Classical Arabic, CS = Code-Switched Arabic, Dia. = diacritic. The table reports the number of sentences in each split.}
\label{tab:dataset}
\end{table}
\subsection{Related Work}
Research on Arabic DR has evolved from rule-based methods to neural and multimodal approaches \cite{elgamal_review2024}. Early systems relied on lexicons and morphological analyzers, later extended with n-gram models \cite{habash2007arabic, elshafei2006automatic}, but they struggled with dialectal variation, noisy text, and borrowed vocabulary. Neural models, from RNNs and LSTMs \cite{zitouni2006arabic, belinkov2015arabic} to transformers \cite{nazih2022arabic} with pre-trained language models such as AraBERT \cite{antoun2021araberttransformerbasedmodelarabic}, CAMeLBERT \cite{inoue2021interplay}, and CATT \cite{alasmary2024catt}, improved accuracy but still failed to resolve phonetic ambiguities in dialects. While \cite{elgamal_review2024} highlighted the usefulness of “diacritics-in-the-wild” signals, text-only models remain insufficient for ambiguous cases.

\textbf{Multimodal} approaches increasingly exploit ASR outputs as phonetic cues. Early work \cite{Aldarmaki_2023} relies solely on ASR, which can produce both transcripts and diacritic predictions, but errors in transcription often propagate to diacritization. More recent methods \cite{shatnawi2024automaticrestorationdiacriticsspeech} integrate ASR-derived-diacritized transcripts with undiacritized text via cross-attention, enhancing performance while still being sensitive to ASR noise.

\textbf{Our approach differs by} (i) deeply integrating text and speech through early and cross-attention fusion, (ii) focusing explicitly on dialectal DR with robust pre-trained encoders: CATT and Whisper.

\section{System Overview}
\label{sec:system}
\begin{figure*}[t]
  \centering
  \begin{subfigure}[c]{0.5\linewidth}  
    \centering
    \includegraphics[width=0.90\linewidth, keepaspectratio]{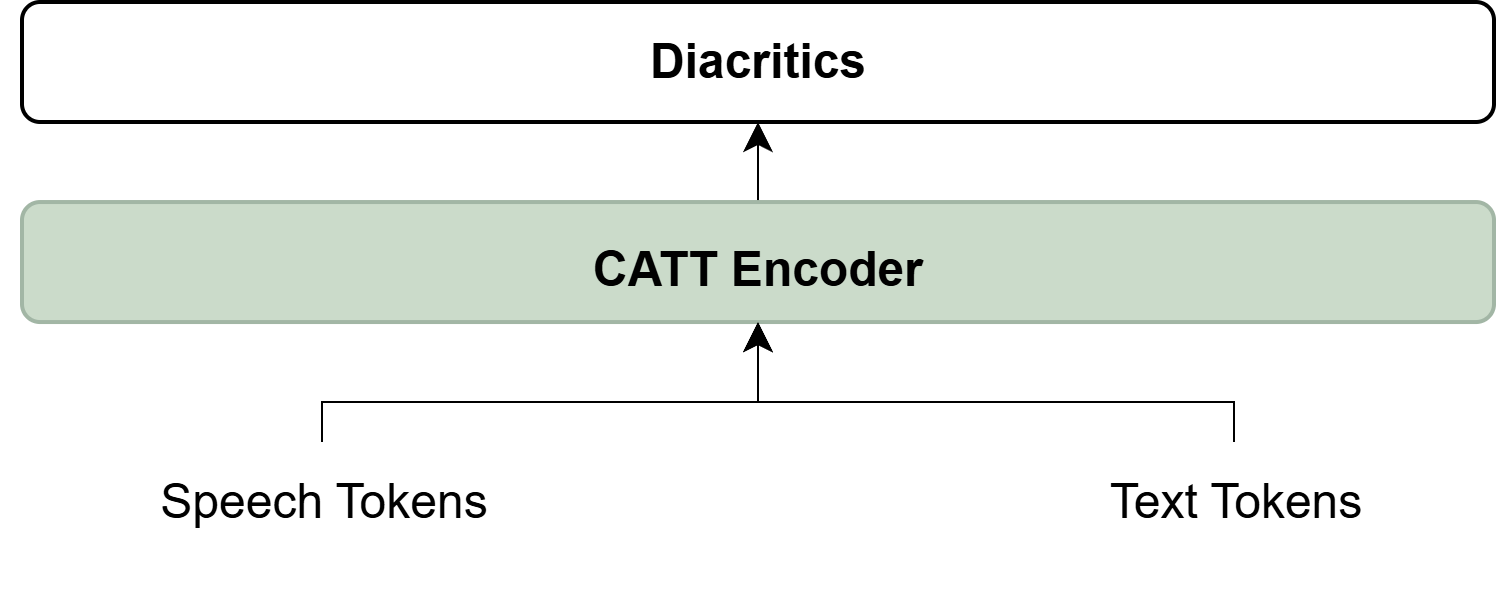}
    \caption{Early Fusion}
    \label{fig:early-fusion}
  \end{subfigure}\hfill
  \begin{subfigure}[c]{0.5\linewidth}  
    \centering
    \includegraphics[width=0.90\linewidth, keepaspectratio]{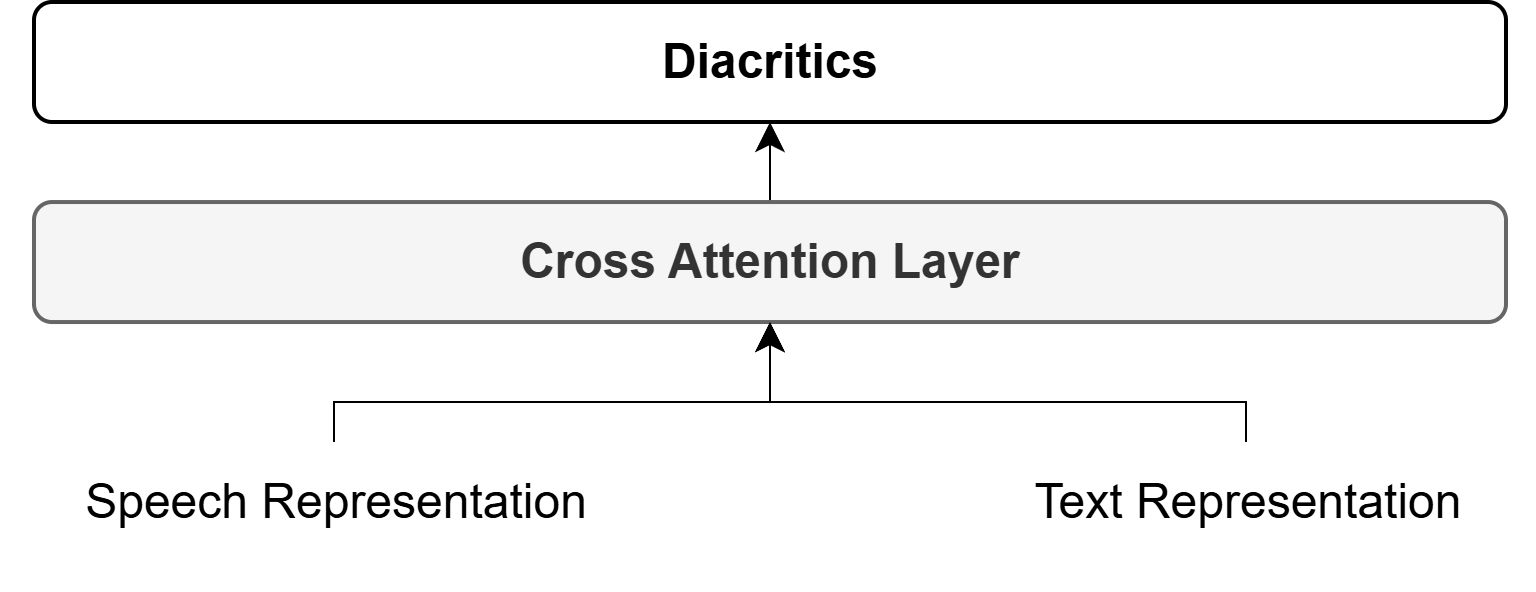}
    \caption{Cross Attention Fusion}
    \label{fig:cross-attention-fusion}
  \end{subfigure}
  \caption{Proposed CATT-Whisper Architectures for Multimodal. (a) Early Fusion Configuration. (b) Cross-Attention Fusion Configuration.}
  \label{fig:two-png-images}
\end{figure*}


\subsection{Architecture Components}



The architecture consists of a \textbf{Text Encoder}, implemented with a pre-trained CATT model for DR, and a \textbf{Speech Encoder}, implemented with the Encoder part of Whisper-Base model. A \textbf{Linear Projection Layer} follows the speech encoder to match the dimensionality of the text encoder. The proposed architectures are summarized in Figure \ref{fig:two-png-images}.

\subsection{Fusion Strategies}
\subsubsection{Early Fusion}

\begin{figure*}[t]
  \centering
  \includegraphics[width=0.90\linewidth, keepaspectratio]{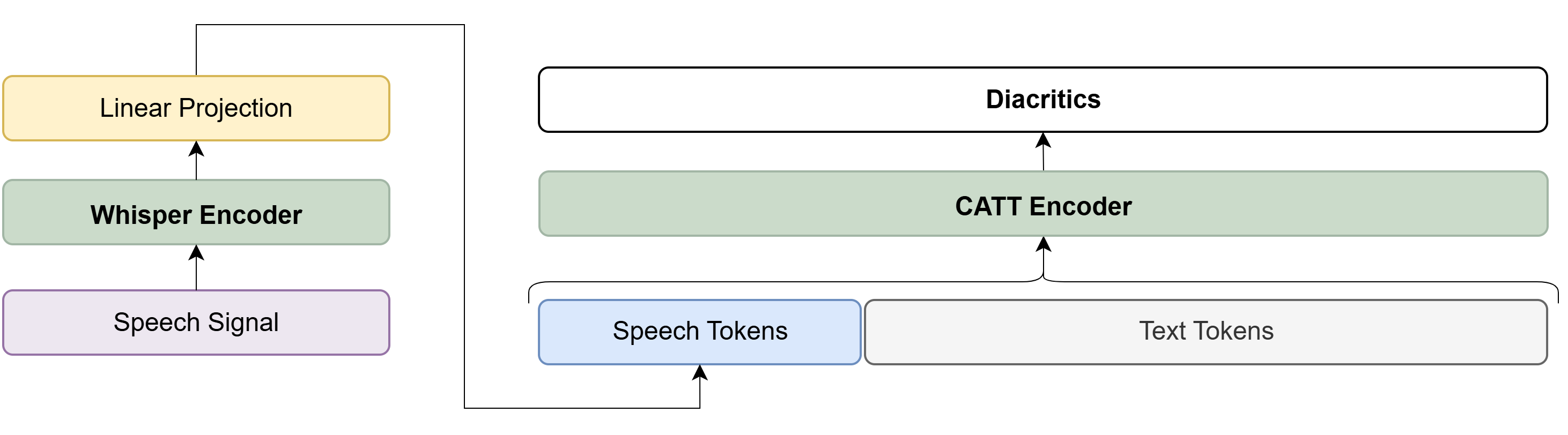}
  \caption{Early Fusion architecture of the proposed CATT-Whisper model. Speech features are downsampled and projected to match text embeddings before being concatenated with text tokens and processed by the CATT encoder.}
  \label{fig:early-fusion-details}
\end{figure*}

Speech features are downsampled from 1,500 frames to 150 tokens by averaging 10 frames with and projecting them to match the text embedding dimension. These speech tokens are then concatenated with text tokens and fed into the CATT encoder, following a strategy similar to \cite{10389705}. This early fusion approach can be seen as a form of “soft prompting,” where text tokens are augmented with speech embeddings via speech-placeholder tokens, enabling the model to leverage acoustic features while preserving the core CATT architecture. Details of this fusion strategy is shown in Figure \ref{fig:early-fusion-details}.

\subsubsection{Cross-Attention Fusion}
Text and speech embeddings are encoded separately, then fused via a cross-attention layer before being passed to the classification layer, similar to the multi-modal setup of \cite{shatnawi-etal-2024-data}.


\subsubsection{Fusion Strategy Choice}
In our experiments, both Early Fusion and Cross-Attention Fusion yielded comparable results. However, as Cross-Attention is computationally more demanding, we focused on Early Fusion, and all results reported in this paper correspond to this configuration.

\subsection{Speech Augmentation}
Time-frequency warping \cite{Park_2019} is applied during training to improve generalization.





\section{Experimental Setup}

For training, we used the NADI 2025 DR train and development sets, while evaluation was performed on the official test set. Model performance was measured using Word Error Rate (WER) and Character Error Rate (CER), which are the standard metrics.  Our preprocessing step included tokenization, speech feature extraction, and spectrogram augmentation through time-frequency warping.

Training was carried out with a batch size of $32$, a learning rate of $1 \times 10^{-5}$, a dropout rate of $0.1$, and the AdamW optimizer.
During training, the speech encoder was frozen for the first 5 epochs allowing the projection layer to adapt, then unfrozen and jointly trained with the rest of the model for more 5 epochs. This two-phase procedure was applied in all experiments for both fusion models.

\section{Results}

\subsection{Development Set Performance}
Table~\ref{tab:dev_results} shows the performance of our proposed model compared to other works on the development set. Our model achieves substantially lower word error and character error rates (WER and CER).

\begin{table}[h]
\centering
\begin{tabular}{lcc}
\hline
\textbf{Participant} & \textbf{WER} & \textbf{CER} \\
\hline
\textbf{gahmed92 (Ours)}      & \textbf{0.25} & \textbf{0.09} \\
omarnj    & 0.46 & 0.22 \\
Baseline  & 0.46 & 0.22 \\
\hline
\end{tabular}
\caption{Results on the NADI 2025 Subtask 3 official development set, reported in WER and CER}
\label{tab:dev_results}
\end{table}

\subsection{Test Set Performance}
Table~\ref{tab:test_results} presents the results on the official test set. Our models outperforms all models in both metrics.

\begin{table}[h]
\centering
\begin{tabular}{lcc}
\hline
\textbf{Participant} & \textbf{WER} & \textbf{CER} \\
\hline
\textbf{gahmed92 (Ours)}      & \textbf{0.55} & \textbf{0.13} \\
mohamed\_elrefai  & 0.64 & 0.15 \\
Baseline  & 0.65 & 0.16 \\
\hline
\end{tabular}
\caption{Results on the NADI 2025 Subtask 3 official test set, reported in WER and CER}
\label{tab:test_results}
\end{table}
\subsection{Performance on Challenging Test Cases}
We further analyzed the model on a set of challenging test cases recorded by our team, where the same word is pronounced differently within the same sentence. The results, summarized in Table~\ref{tab:challenging_results}, show that while our model achieves lower WER and CER than the others, these cases remain difficult and are not fully solved. This highlights both the robustness of our approach and the need for further improvements to handle complex, real-world pronunciation variability.
\begin{table}[h]
\centering
\begin{tabularx}{\linewidth}{|l|>{\centering\arraybackslash}X|}
\hline
Example Input 1         &            \<ضرب ضرب ضرب>      \\ \hline
CATT-Whisper            &            \<ضَرِب ضَرِب ضَرِب>     \\ 
Reference               &            \<ضَرَبَ ضُرِبَ ضَرْبٌ>     \\  
\hline
\hline
Example Input 2         &            \<ذهب ذهب>      \\ \hline
CATT-Whisper            &            \<ذَهِب ذَهِب>     \\ 
Reference               &            \<ذَهَبٌ ذَهَبْ>     \\  
\hline
\end{tabularx}
\caption{Model performance on challenging test cases with variable word pronunciations.}
\label{tab:challenging_results}
\end{table}
\section{Conclusion}
We present CATT-Whisper, a multimodal system for Arabic DR that combines pre-trained text and speech encoders via early fusion and cross-attention. Both strategies achieve competitive results. While speech input boosts diacritic accuracy, some ambiguous sequences remain challenging, suggesting the need for stronger phoneme-level encoders (e.g., CTC-based models such as Conformer-CTC \cite{gulati2020conformerconvolutionaugmentedtransformerspeech}, Squeezeformer \cite{kim2022squeezeformerefficienttransformerautomatic}). Future work will explore alternative acoustic models and larger-scale training.

\section*{Acknowledgments}
The authors would like to thank the management of AbjadAI Company for their generous support.

\bibliography{custom}

\end{document}